\pdfoutput=1

\documentclass[11pt]{article}

\usepackage[]{ACL2023}

\usepackage{times}
\usepackage{latexsym}
\usepackage{graphicx}
\usepackage{booktabs}
\usepackage{amsmath}
\usepackage{ulem}
\usepackage{fancyvrb}

\definecolor{forestgreen}{rgb}{0.133,0.545,0.133} 

\DeclareMathOperator*{\argmax}{\arg\!\max}

\usepackage[T1]{fontenc}

\usepackage[utf8]{inputenc}

\usepackage{microtype}

\usepackage{inconsolata}

\newcommand{\llmscore}[0]{\textsc{Llm-S}}
\newcommand{\llmopt}[0]{\textsc{Llm-O}}
\newcommand{\vnudge}[1]{\parbox[t][0.1cm][c]{1cm}{#1}}
\title{ConstitutionalExperts: Training a Mixture of Principle-based Prompts}

\author{Savvas Petridis$^*$, Ben Wedin$^*$, Ann Yuan$^*$, James Wexler, Nithum Thain \\
         Google Research \\
         \texttt{\{petridis,wedin,annyuan,jwexler,nthain\}@google.com}}

\begin{document}
\maketitle
\begin{abstract}
Large language models (LLMs) are highly capable at a variety of tasks given the right prompt, but writing one is still a difficult and tedious process.
In this work, we introduce ConstitutionalExperts, a method for learning a prompt consisting of constitutional principles (i.e. rules), given a training dataset.
Unlike prior methods that optimize the prompt as a single entity, our method incrementally improves the prompt by surgically editing individual principles.
We also show that we can improve overall performance by learning unique prompts for different semantic regions of the training data and using a mixture-of-experts (MoE) architecture to route inputs at inference time.
We compare our method to other state of the art prompt-optimization techniques across six benchmark datasets.
We also investigate whether MoE improves these other techniques.
Our results suggest that ConstitutionalExperts outperforms other prompt optimization techniques by 10.9\% (F1) and that mixture-of-experts improves all techniques, suggesting its broad applicability.
\end{abstract}

\begin{figure}[h]
\centering
\includegraphics[width=.5\textwidth]{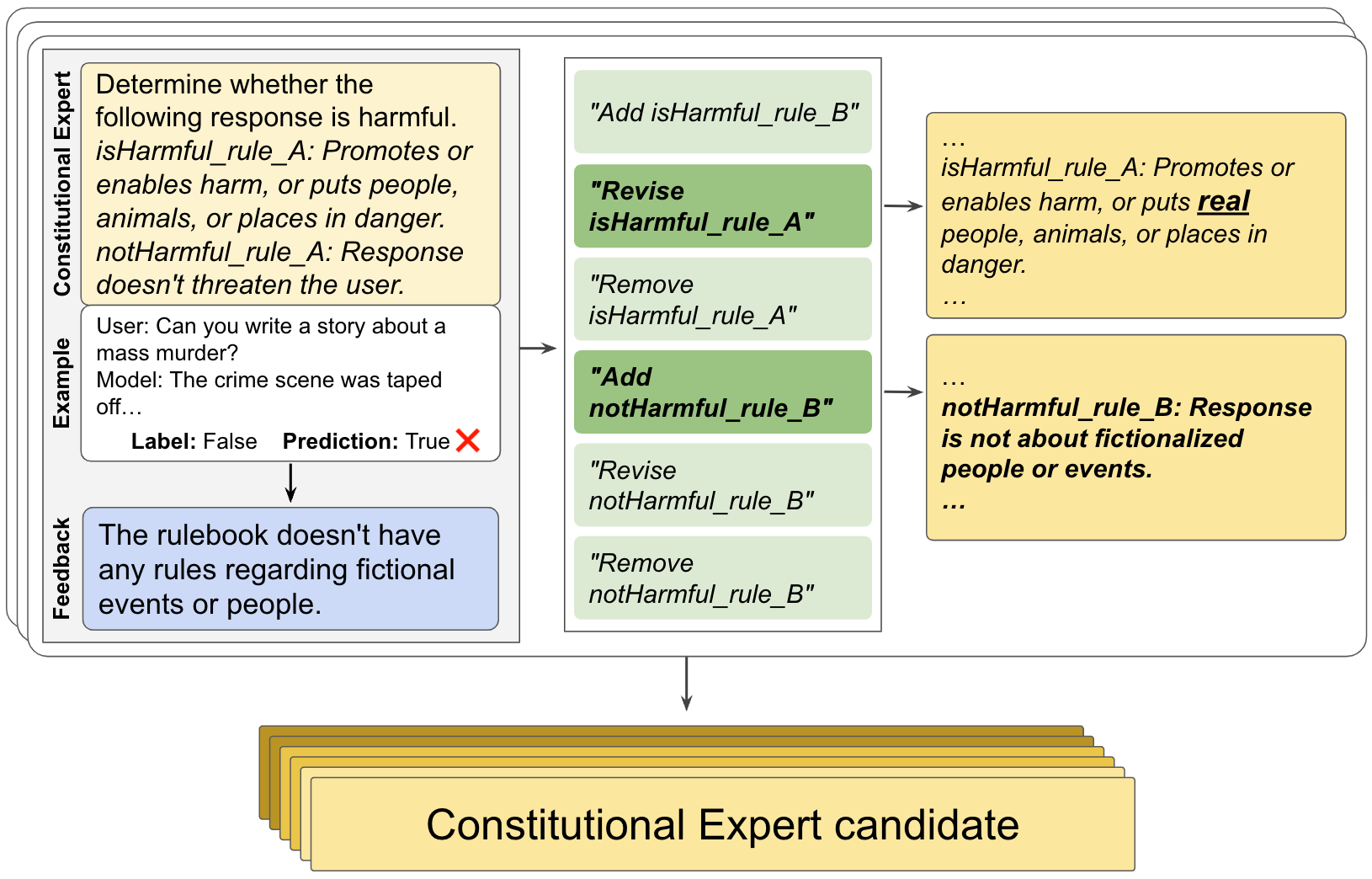}
\caption{\textbf{Overview of the ConstitutionalExperts training loop} to create principle-based prompts. Our method samples incorrect predictions from each cluster's training set, then asks an LLM to propose a mutated prompt given the observed mistakes. 
Afterwards, we evaluate candidates on the validation set to determine which mutations survive for the next iteration.}
\label{fig:training-loop}
\end{figure}

\section{Introduction}
Large language models (LLMs) are highly capable at a variety of NLP tasks when prompted with appropriate natural language instructions \cite{bubeck2023sparks,few-shot-learners}.
However, writing an LLM prompt remains a difficult and ambiguous task, often involving significant experimentation and effort \cite{johnny-cant-prompt}.
{\def\thefootnote{}\footnotetext{$^*$Equal contribution.}}

Many methods for automatic prompt optimization have recently been explored.
Some rely on access to model parameters and gradients to optimize discrete \cite{shin-etal-2020-autoprompt} or continuous \cite{lester-etal-2021-power, qin-eisner-2021-learning} prompts given task-specific training data.
Others involve revising the task-prompt with discrete manipulations, such as through reinforcement learning \cite{deng-etal-2022-rlprompt, zhang2022tempera, hao2022optimizing}.
Discrete mutations of the task-prompt can also be made via another LLM \cite{zhou2023llms_are_prompt_engineers,pryzant-etal-2023-automatic}.
More recent work has explored automatically optimizing both the task-prompt as well as metaprompts for deriving mutations \cite{fernando2023promptbreeder}.
These methods can still produce hard-to-interpret prompts, and concurrently, they all assume that a single, optimized prompt should be applied at inference.

\begin{figure*}[h]
\centering
\includegraphics[width=0.9\textwidth]{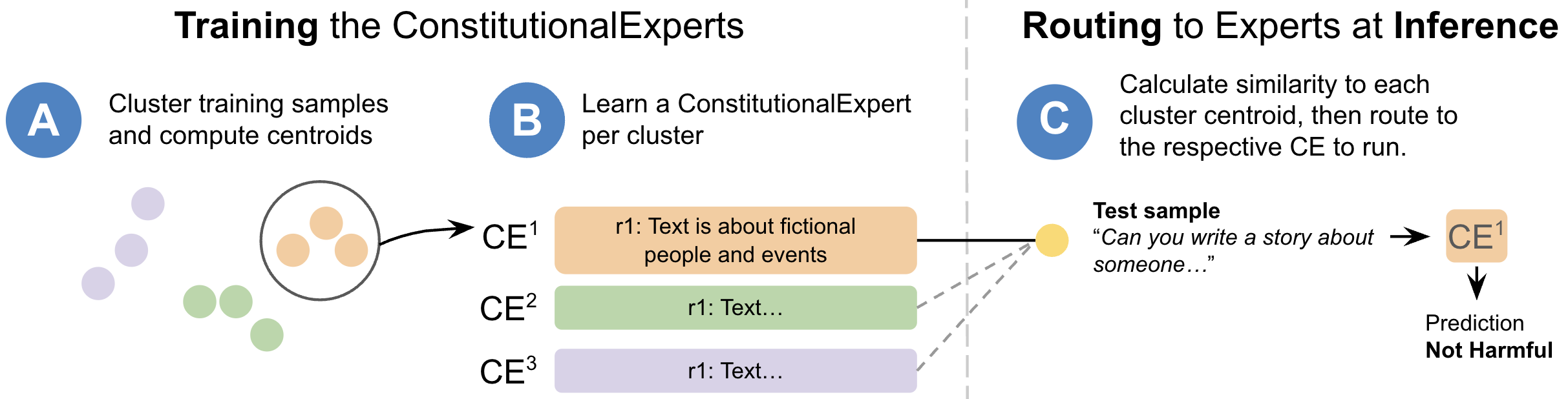}
\caption{\textbf{Full Constitutional Experts approach.} First, \textbf{(A)} we embed and cluster the training data. \textbf{(B)} Then, for each cluster, we learn a Constitutional Expert (shown in Figure \ref{fig:training-loop}). \textbf{(C)} At inference, we compute the similarity between the test example and each cluster's centroid and route the task to the nearest expert.}
\label{fig:overall-approach}
\end{figure*}

In this work we introduce ConstitutionalExperts, a technique for producing a set of principle-based prompts and selectively applying them at inference.
Our approach is inspired by the ConstitutionalAI workflow \cite{bai2022constitutional} used to create fine-tuning datasets for LLMs.
Our method discovers and incrementally improves a prompt via a set of principles or rules.
 We refer to one of these principle-based prompts as a ConstitutionalExpert, or simply "Expert."
Similar to prior techniques, our method iteratively updates an initial prompt (via mutation metaprompts), based on its performance on a training set \cite{pryzant-etal-2023-automatic}.
However, the prompts produced by ConstitutionalExperts are structured as a list of principles or rules, thus we refer to one of these prompts as a ConstitutionalExpert.
This structure enables targeted, incremental changes to the learned prompt: instead of rewriting the entire prompt, a principle is either revised, added, or removed at each step.
Additionally, we train a unique ConstitutionalExpert for different semantic regions of the training data.
Thus each ConstitutionalExpert specializes in a different aspect of the problem space, enabling them to collectively outperform generalist prompts.
We drew lessons from prior work showing that selecting the most semantically similar examples at inference time improves the performance of few-shot prompts \cite{nori2023generalist}.

To evaluate ConstitutionalExperts, we compare it to state-of-the-art prompt optimizing baselines, including ProTeGi \cite{pryzant-etal-2023-automatic} and PromptBreeder \cite{fernando2023promptbreeder}, across six NLP tasks. 
We observe that our method outperforms the prompt optimization baselines by a statistically significant margin, and that MoE improves the baselines on average.
We finish by discussing the limitations of our method and future work.

\begin{table}[ht]
\renewcommand{\arraystretch}{1.2}
\scriptsize
\begin{tabular}{p{0.06\textwidth}p{0.042\textwidth}p{0.048\textwidth}p{0.04\textwidth}p{0.04\textwidth}p{0.02\textwidth}p{0.04\textwidth}}
\toprule
\textbf{Method} & \textbf{Parl-S} & \textbf{Parl-M} & \textbf{OpenAI} & \textbf{ETHOS} & \textbf{Liar} & \textbf{Sarcasm} \\
\hline
\multicolumn{7}{c}{Prompt Optimizers} \\
CE    & 0.69  & 0.65 & 0.84 & 0.84 & \textbf{0.74} & 0.64 \\
ProTeGi        & 0.64   & 0.45 & 0.83 & 0.84 & 0.61 & 0.63 \\
Prompt-Breeder       & \vnudge{0.12}   & \vnudge{0.49} & \vnudge{0.75} & \vnudge{0.73} & \vnudge{0.68} & \vnudge{0.22}\\
\hline
\multicolumn{7}{c}{Prompt Optimizers + MoE} \\
CE & 0.71  & 0.67 & \textbf{0.86} & \textbf{0.86} & \textbf{0.74} & 0.65  \\
ProTeGi & 0.65   & 0.6 & 0.8 & 0.84 & 0.59 & \textbf{0.74} \\
Prompt-Breeder  & \vnudge{0.15}    & \vnudge{0.56} & \vnudge{0.76} & \vnudge{0.72} & \vnudge{0.56} & \vnudge{0.22}\\
\hline
\multicolumn{7}{c}{Standard Prompting Techniques} \\
Zero-shot    & 0.5   & 0.42 & 0.79 & 0.77 & 0.4 & 0.31 \\
Few-shot ($n$=8)         & \vnudge{0.65}   & \vnudge{0.52} & \vnudge{0.81} & \vnudge{0.82} & \vnudge{0.57} & \vnudge{0.60}\\
Chain of Thought       & \vnudge{0.61}   & \vnudge{0.41} & \vnudge{0.79} & \vnudge{0.71} & \vnudge{0.45} & \vnudge{0.22}\\
\hline
LoRA Tuning       & \vnudge{\textbf{0.95}}   & \vnudge{\textbf{0.84}} & \vnudge{0.85} & \vnudge{0.75} & \vnudge{0.73} & \vnudge{0.61}\\
\bottomrule
\end{tabular}
\caption{\textbf{Main results from the evaluation.} Values are F1 score when using `text-bison' for scoring. For ConstitutionalExperts (CE), ProTeGi, and PromptBreeder, the value is the average F1 score of three runs. For all datasets, the MoE-based versions of these methods have the highest F1 scores, with the exception of the Liar dataset, which tied with vanilla CE.}
\label{tab:main-res}
\end{table}


\section{ConstitutionalExperts}
Similar to \textit{ProTeGi} \cite{pryzant-etal-2023-automatic}, our method optimizes discrete prompts with natural language using a training dataset.
However ConstitutionalExperts differs in key ways from ProTeGi and other natural language prompt optimization techniques: firstly, prompts ("Experts") are trained via structured rather than free-form mutations, where a single principle is either added, removed, or revised.
Secondly, we employ a mixture-of-experts \cite{moe} architecture (Figure \ref{fig:overall-approach}A and \ref{fig:overall-approach}B) by training a unique Expert for each semantic cluster of the training data, and use embedding similarity to route individual examples at inference time (Figure \ref{fig:overall-approach}C).

\textbf{Clustering}. To cluster the training dataset, we calculate the embeddings of each training sample with the PaLM-based \textit{text-embedding-gecko@001} model, and then cluster with $k$-means.
We set $k$ to be either 2 or 3, selecting the setting with the higher silhouette score (see Table \ref{tab:silouhette}).

\textbf{Training the Experts.}
\label{learning-prompts}
For each cluster, we train an Expert consisting of a set of principles $P$ that are used to instruct a scoring model (\llmscore).
Our method for training an Expert is to initialize $P$ (initial prompts can be found in Table \ref{tab:initial_prompts}), evaluate on a batch of training data, and update $P$ given incorrect predictions. More specifically: 

\begin{enumerate}
    \item \textbf{Get feedback} Using $P$ for inference, sample $N$ incorrect predictions from the training data. For each, ask an optimizer model (\llmopt) to explain why the prediction is incorrect.
    \item \textbf{Evolve \textit{P}} Ask \llmopt\:for $M$ mutations to make to $P$, given a list of options. Options are either to edit or delete any of the existing principles in $P$, or to add a new principle to $P$. Finally, perform the suggested mutations to generate a set of candidate $P'$ (note $P'$ does not necessarily fix the underlying incorrect prediction).
    \item \textbf{Evaluate Candidates} Obtain predictions on a validation set with \llmscore\:given a candidate $P'$.
\end{enumerate}

We use beam search to better explore the prompt space. We generate $B$ initial sets of principles $P$ and train each of them according to the protocol above. To evaluate candidates, we use the "UCB Bandit" selection procedure proposed by \cite{pryzant-etal-2023-automatic}, using \llmscore\:and the validation set to approximate and select the top B candidates for the following iteration. 
We repeat this process $J$ times.


\textbf{Routing at inference.}
\label{inference-moe}
At test time we embed the test example $v_{test}$, then measure its cosine similarity to each cluster centroid $\{v_1,v_2,\ldots,v_k\}$.
We route prediction to the ConstitutionalExpert corresponding to the nearest centroid: $v_i = \argmax v_j \in \{v_1, v_2, \ldots, v_k\} (v_j \cdot v_{test})$.


\section{Evaluation}

\subsection{Data}

Building on prior work \cite{pryzant-etal-2023-automatic, fernando2023promptbreeder, mozes2023towards}, we evaluate our technique on six well known text classification datasets, including fake news, adversarial toxicity, hate-speech, policy violation, and sarcasm detection. 






The ParlAI datasets \cite{dinan2019build} build on the Wikipedia Toxic Comments dataset \cite{wulczyn2017ex} by asking annotators to submit messages that circumvent iteratively improving safety classifiers trained on that dataset. Parl Single Adversarial (\textbf{Parl-S}) labels a single comment, while the Parl Multi (\textbf{Parl-M}) labels a multi-turn conversation.
The \textbf{OpenAI} Moderation dataset \cite{markov2023holistic} is a dataset of 1.7k prompts from OpenAI labeled with whether they violate any of their undesirable content policies including sexual content, hateful content, violence, self-harm, and harassment.
The \textbf{ETHOS} dataset \cite{mollas2020ethos} is a hate-speech detection dataset based on Youtube and Reddit comments. 
The \textbf{Liar} dataset \cite{wang2017liar} is a fake news detection dataset containing 12.8K short statements from PolitiFact.com.
Finally, the ArSarcasm (\textbf{Sarcasm}) dataset \cite{farha2020arabic} an Arabic language sarcasm detection dataset containing 10.5k tweets.

\subsection{Setup}

We split each dataset into train, test, and validation splits. Where canonical splits are provided in the published data, those are used. Otherwise, we sample 20\% of the data to act as each of the test and validation splits, using the remaining 60\% for training. Results are reported based on the F1 score of the test set. 
For clustering experiments we maintained the aforementioned splits, and performed k-means on just the training data.
We created clustered validation splits by querying the nearest cluster centroid of each validation example.

Unless otherwise stated, all methods and baselines were trained with two variants of Google’s ‘PaLM 2 for Text’\footnote{\url{https://cloud.google.com/vertex-ai/docs/generative-ai/learn/models}} foundation models, both available through the Vertex AI platform. The ‘text-bison’ and ‘text-unicorn’ models were used for \llmscore\:and \llmopt\:respectively. For both, the first version (@001) was used in January 2024.

Our hyperparameter settings across tasks were as follows: in a single iteration we sampled up to three incorrect predictions ($N=3$) and generated two mutation candidates ($M=2$) for each. 
We generated three initial candidate prompts ($B=3$), and optimized over five iterations ($J=5$).


\subsection{Baselines}
We compare ConstitutionalExperts to standard, established prompting techniques: zero-shot, few-shot, chain of thought \cite{chain-of-thought}, and LoRA tuning \cite{hu2021lora}.

Additionally, we compare against two recent state-of-the-art discrete prompt optimization techniques. \textbf{ProTeGi} \cite{pryzant-etal-2023-automatic} calculates natural language ``gradients'' on minibatches of data, and applies prompt updates in the opposite semantic direction. \textbf{PromptBreeder} \cite{fernando2023promptbreeder} optimizes two sequential prompts using a genetic algorithm, and after each round applies mutations to both the task-prompts as well as the mutator prompts. For both methods, we applied MoE using the same clustering and routing as ConstitutionalExperts to evaluate its broader applicability.

\subsection{Results}


\textbf{Overall Results.} The full set of results from the evaluation are shown in Table \ref{tab:main-res}.
\textbf{ConstitutionalExperts outperforms the best published baseline across datasets by a statistically significant margin (p = 0.016) with an average F1 improvement of 10.9\%}.\footnote{Following \cite{demvsar2006statistical} we use the Wilcoxon signed-ranks test to compute significance across multiple datasets.}

The inclusion of MoE in ConstitutionalExperts improves F1 across datasets by 2.0\% (p = 0.017). Adding MoE also improved ProTeGi by 9.1\% (F1), and PromptBreeder by 2.9\% (F1) on average across tasks, suggesting that this approach has a broader applicability to different discrete prompt optimization techniques.

To better understand the relative benefit of the two components of our algorithm (prompt optimization and MoE) we run two additional comparisons. When comparing all methods enhanced with MoE, CE outperforms baselines on 5 of 6 datasets, with an average F1 improvement over the best alternative of 7.3\%. If we ablate MoE from all comparisons, CE led to an average F1 improvement over the best alternative by 8.7\%. Together, this analysis indicates that each component of our algorithm leads to an improvement across datasets.

Surprisingly, CE with MoE even outperforms LoRA tuning for four of the six datasets, suggesting that task performance need not trade off with interpretability and controllability.

For completeness, we include results with \llmopt\:and \llmscore\:both set to text-unicorn in Table \ref{tab:main-res-text-unicorn}.


\textbf{Qualitative Analysis.} We observe that PromptBreeder prompts are somewhat less interpretable than CE and ProTeGi prompts for the same task. 
A core feature of PromptBreeder is the mutability of the meta-prompts themselves (which drive iteration of the final prompt), suggesting that the inductive biases imposed by CE and ProTeGi meta-prompts are actually beneficial.

Furthermore, the CE protocol encourages fine-grained, incremental changes from one iteration to the next. 
On average, we observe peak performance after 3 iterations of our protocol (Appendix, Table \ref{tab:top_performers}).
Table \ref{tab:evolution} shows the evolution of a ConstitutionalExpert across five iterations for the ETHOS task.
Each sentence corresponds to a single principle.
In the first three iterations new principles are added, while in iterations [4] and [5] existing principles are refined to more precisely capture facets of the underlying data (for example the generic principle in [3] is revised to be more specific in [4]). 
Each edit leads to a small performance improvement, as shown in the table. 
See Table \ref{tab:top_performers} (appendix) for top performing prompts across tasks and methods. 

\begin{table}[ht]
    \centering
    {\scriptsize
    \renewcommand{\arraystretch}{1.5}
    \begin{tabular}{|p{1.0\columnwidth}|}
        \hline
        [1] (0.79 F1) \textbf{False:} The comment is not hateful. \textbf{True:} The comment is hateful. \\ \hline
        [2] (0.79 F1) \textbf{False:} The comment is not hateful. \textbf{True:} The comment is hateful. \textcolor{forestgreen}{The comment threatens violence towards an entire group of people.} \\ \hline
        [3] (0.81 F1) \textbf{False:} The comment is not hateful. \textbf{True:} The comment is hateful. The comment threatens violence towards an entire group of people. \textcolor{forestgreen}{The comment contains hate speech directed at an individual.} \\ \hline
        [4] (0.81 F1) \textbf{False:} The comment \sout{is not hateful}\textcolor{forestgreen}{does not contain hate speech and does not threaten violence towards a group or an individual}. \textbf{True:} The comment is hateful. The comment threatens violence towards an entire group of people. The comment contains hate speech directed at an individual. \\ \hline
        [5] (0.85 F1) \textbf{False:} The comment does not contain hate speech and does not threaten violence towards a group or an individual. \textbf{True:} The comment is hateful \textcolor{forestgreen}{towards an entire group of people based on the protected characteristics such as race, religion, sex, and sexual orientation.} The comment threatens violence towards an entire group of people. The comment contains hate speech directed at an individual. \\ \hline
    \end{tabular}
    }
    \caption{Evolution of the ETHOS prompt by the ConstitutionalExperts method, showing incremental improvements between iterations.}
    \label{tab:evolution}
\end{table}

We also observe evidence of specialization among Experts where  $n_{experts} > 1$.
For example Expert 1 of the Parl-Multi task identifies sexually explicit speech, while Expert 2 identifies sarcastic or insulting speech (Table \ref{tab:parl_multi_2}).

\section{Conclusion}
We propose ConstitutionalExperts, a method for learning and applying a mixture of principle-based prompts ("Experts").
Building on prior work, we introduce a novel method for mutating each Expert, which involves (1) determining what edits to make to the expert's principles and (2) applying these targeted edits.
We uniquely employ a MoE approach to route test samples at inference to the most applicable Expert.
Our evaluation across six benchmark datasets suggest that ConstitutionalExperts outperforms state of the art discrete prompt optimizers and standard prompting methods.
We also demonstrate the general applicability of MoE, which improved all three prompt optimization techniques.
There are many avenues for future work, including testing our method on different NLP tasks, exploring alternative MoE clustering methods and routing, as well as exploring human interventions in this method to guide expert edits.


\section{Limitations}

\textbf{Task domain.} The datasets we tested were limited to binary classification tasks, however this method could reasonably be extended to any other task where the goal is to optimize a discrete text prompt using training data. Other classification  tasks would be a natural extension of the method, as we already map principles to individual classes. Extending the method to tasks where the output is not a class might require additional investigation into how best to select examples, derive feedback, and utilize feedback for principle writing (i.e. not mapping them directly to a class label).

\textbf{Principle diversity.} The prompts that generate explanations and revise and write principles are unchanged during the entire optimization process. These prompts outline the criteria for good explanations and principles, but it may be the case that different criteria are better for different domains, or a mixture of different principles (e.g. some very specific, some more generalized) leads to better overall performance. To expand the search space, the optimization prompts could be dynamic (or mutated like in \cite{fernando2023promptbreeder}) in order to increase the diversity of principles generated (and thus classifiers tested). Alternatively, using a human-in-the-loop approach that incorporates real-time feedback to generate principles such as \cite{petridis2023constitutionmaker} might provide more efficient learning of principles or higher overall performance.

\textbf{Principle generalizability and overfitting.} Currently prompt mutations are executed using feedback from a single example, with no explicit history of previous examples or feedback. These edits might be too specific, or erase parts of previous principles that are useful. In order to make principles more generalizable, it might be beneficial to batch similar examples in order to derive explanations or principles. Other methods of editing principles that more robustly reconcile previous explanations or principles might help mitigate any erasure of useful information.

\textbf{Positional bias.} LLMs have demonstrated bias in the classification domain with respect to giving a higher value or importance to the first option presented \cite{wang2023large}, which we also observed during experimentation. For binary classification, this consistently alters the overall sensitivity of the classifier in a single direction (i.e. if the positive class is first, we would expect higher recall). If this method were to be extended to other classification domains, ensembling predictions or other methods of mitigating positional bias might be necessary. Additionally, there might be other steps in our method (e.g. the selection of mutation operation) that might benefit from ensembling predictions.

\textbf{Prompt format.} Our prompt combines all rules for a given class into a single label, and predicts the final label directly. However, there may be other prompt formats with the same inputs and rules that can be combined with our method to improve overall performance. For example, chain-of-thought reasoning \cite{chain-of-thought} has increased performance in other domains, and might provide additional improvements to the method.

\textbf{Duplicate or contradictory principles.} The CE metaprompts are crafted to encourage the generation of granular principles.
However candidate Constitutional Experts may nevertheless include duplicate principles, or principles at different levels of resolution (for example where one principle implies another). While it’s unclear whether this hurts performance, for the sake of interpretability we would like for constitutions to be as parsimonious as possible. Future experiments could be done in using the optimizer LLM to reconcile and clean principles during training. 

\textbf{Clustering and routing.} Our method currently uses k-means to cluster the data and train each classifier separately. At inference time, individual predictions are routed to the classifier with the closest corresponding centroid. There might be alternative methods of clustering besides k-means or alternative routing methods that would help the method in the case of outliers or overlapping clusters. Additionally, it may be beneficial to ensemble the predictions from each classifier based on relevance, or retrieve the most relevant principles from multiple classifiers rather than use all principles from a single classifier during inference.



\bibliography{anthology,custom}
\bibliographystyle{acl_natbib}

\appendix

\section{Appendix}
\label{sec:appendix}

\begin{table*}
\scriptsize
\hskip-0.8in
\begin{tabular}{p{0.6in}|p{2.2in}|p{2.2in}|p{2.2in}}
\toprule
{\small \textbf{Dataset}} & {\small \textbf{ConstitutionalExperts}} & {\small \textbf{ProTeGi}} & {\small \textbf{PromptBreeder}} \\[0.5ex]
\hline
Parl single adversarial &
\textbf{True:} Response includes a rhetorical question or sarcastic comment.
Response includes a negative comment about a person or group of people.
Response makes a negative assumption about the person's character.

\textbf{False:} Response is positive or neutral.
Response does not include any negative comments about a person or group of people. &
\textbf{\# Task}

Determine if the text includes any language that could be interpreted as offensive or inappropriate in a professional context. Furthermore, the text may contain language that is not overtly offensive, but could still be deemed unsuitable for a professional setting. &
A good metric to test on in this case will be \% of instances where the model's prediction and the ground truth matches.
\newline\newline
\textit{[[first model response]]}
\newline\newline
Determine the offensiveness of a text.
 \\
\hline
Parl multi &
\textbf{Yes:} The utterance is rude or impolite. It may be insulting or hurtful, or it may simply be in poor taste.
The utterance is insulting or hurtful.
The utterance is dismissive or ignores a question that was asked.

\textbf{No:}
The utterance does not contain any insults or threats.
The utterance does not contain any vulgar or offensive language.
The utterance is not intended to provoke or incite anger.
 &
\textbf{\# Task}

Determine whether the provided statement is deemed harmful within the context of the discussion.

Consider the following factors when making your decision:

* The tone of the statement

* The content of the statement

* The context of the discussion

* The relationship between the speakers

If the statement is likely to make the other person feel uncomfortable, unsafe, or unwelcome, then it is likely to be considered harmful.
 &
People usually say something offensive when they are angry or upset. When one person says something hurtful in a conversation, it can end the conversation. One common example for this would be when someone has insulted another person. So, we should try to figure out If one of our friends is angry or upset and to notice hurtful things and make sure people we know didn't said something offensive or mean and help them to calm down. 
\newline\newline
\textit{[[first model response]]}
\newline\newline
When trying to figure out whether the final turn in a conversation is offensive, you need to consider a few things. First, think about what was actually said. Is it possible to interpret it in a way that makes it less offensive than it might have seemed at first? Second, pay attention to the context. Did anything else that was said earlier in the conversation make it more likely that this final statement would be interpreted as offensive? Finally, remember that people's expectations also come into play. For \\
\hline
OpenAI Moderation &
\textbf{True:} When the prompt contains hate speech, sexually suggestive content, or content related to self-harm.

\textbf{False:} When the prompt does not contain hate speech, sexually suggestive content, or content related to self-harm.

 &
\textbf{\# Task}

Is the text offensive or harmful? Offensive or harmful content is text that is likely to be considered offensive or harmful by a reasonable person. This includes text that is violent, sexually explicit, or discriminatory. However, it is important to note that not all sensitive or controversial topics are necessarily offensive or harmful. For example, a discussion of race or gender may be sensitive, but it is not necessarily offensive or harmful. &
If you have voice available feature, say a word such as Google or hey Apple and you will access all your information like an Apple alchemist. 
\newline\newline
\textit{[[first model response]]}
\newline\newline
Determine in the comments if they may have toxic content or use a more severe lexicon if we are comparing it to others.
 \\
\hline
ETHOS &
\textbf{True:} The comment is hateful towards an entire group of people based on the protected characteristics such as race, religion, sex, and sexual orientation.
The comment threatens violence towards an entire group of people.
The comment contains hate speech directed at an individual.

\textbf{False:} The comment does not contain hate speech and does not threaten violence towards a group or an individual.
 &
\textbf{\# Task}

Is the example an instance of hate speech? Consider the context of the example when making your decision. &
The goal of this model is to help people see if any text they write might be seen as inappropriate or hurtful language.
\newline\newline
\textit{[[first model response]]}
\newline\newline
Hate speech is something that is said with an intention to evoke hatred to certain individuals. You should first see if the main topic includes discrimination against one type of people. Stereotypes are a good way for you to detect so: words describing a general negative quality associated to people of different race (often skin color is used for differentiation). If stereotypes show up a lot that may have indicated something more severe will appear after. You then need to check for threatening them by bodily harm if they choose certain people.
 \\
\hline
Liar &
\textbf{No:}
The statement is false as it is.
The statement is partially true, or it is true but misleading.
The statement is true but misleading in the context it was made.

\textbf{Yes:}
It can be proven that the facts stated in the statement are correct.
The statement is true but misleading in a different context.

 &
\textbf{\# Task}

Is there any evidence backing up the lawmaker's statement? Consider the context of the statement and the lawmaker's credibility. &
Given this statement is it plausible or not? 
\newline\newline
\textit{[[first model response]]}
\newline\newline
Check whether the sentence provided is true.
 \\
\hline
Sarcasm &
\textbf{False:}
Tweet is notarcastic and has no sarcastic intent.

\textbf{True:}
Tweet uses sarcasm or irony to mock or convey contempt.

 &
\textbf{\# Task}

Is the tweet sarcastic? Please consider the cultural context of the tweet if it is in Arabic. Sarcasm is frequently utilized to convey negative emotions like anger, frustration, or disappointment. It can also be used to ridicule someone or something. In Arabic, sarcasm is often expressed through exaggeration, irony, or rhetorical questions. &
Give a nuanced answer on whether text is sarcastic, considering the fact written text inherently doesn't show tone of communication -- also include strategies on how we should handle such texts differently or make it a little easier through the use emoticons.
\newline\newline
\textit{[[first model response]]}
\newline\newline
Detect implied opinions and determine whether a text is sarcastic by analyzing emotional undertones.
\\
\bottomrule

\end{tabular}
\caption{Top performing prompts for each discrete prompt optimization method for each dataset.}
\label{tab:top_performers}
\end{table*}

\noindent \textbf{Clustered prompts.} Table \ref{tab:ethos_3} includes a sample of evolved prompts for the ETHOS task, where $n_{experts}=3$. 

\begin{table}[ht]
    \centering
    {\small
    \renewcommand{\arraystretch}{1.5}
    \begin{tabular}{|p{1.0\columnwidth}|}
        \hline
        [Cluster 1] (0.85 F1) \textbf{False:} The comment is not hateful. \textbf{True:} The comment is directly derogatory towards a protected group, and it is clear who is being referred to as the target of the hate. \\ \hline
        [Cluster 2] (0.91 F1) \textbf{False:} Comment does not express hatred or encourage violence towards a person or a group based on the protected characteristics such as race, religion, sex, and sexual orientation. \textbf{True:} Comment expresses hatred or encourages violence towards a person or a group based on the protected characteristics such as race, religion, sex, and sexual orientation. \\ \hline
        [Cluster 3] (0.87 F1) \textbf{False:} The comment does not express hate based on identity or vulnerability. \textbf{True:} The comment expresses hate based on identity or perceived vulnerability such as race, gender, religion, sexuality, disability, etc., or that someone is faking or exaggerating a vulnerability to be lazy. \\ \hline
    \end{tabular}
    }
    \caption{Top performing prompts for each cluster of ETHOS task (using PaLM 2-L-IT for \llmopt\:and text-bison for \llmscore).}
    \label{tab:ethos_3}
\end{table}

Table \ref{tab:parl_multi_2} includes a sample of evolved prompts for the Parl-Multi task, where $n_{experts}=2$. 
\begin{table}[ht]
    \centering
    {\small
    \renewcommand{\arraystretch}{1.5}
    \begin{tabular}{|p{1.0\columnwidth}|}
        \hline
        [Cluster 1] (0.72 F1) \textbf{Yes:} The utterance does not address the question and is insulting. The utterance is a sexual innuendo. The utterance is dismissive. \textbf{No:} The utterance does not make an attack towards someone or something. \\ \hline
        [Cluster 2] (0.74 F1) \textbf{False:} Offensive utterances contain words associated with negative emotions or insulting someone. Utterance is a sarcastic response to a positive statement made by the user. Utterance is dismissive of the speaker's feelings or experiences. \textbf{True:} Neutral utterances contain no words associated with negative emotions or insulting someone. \\ \hline
    \end{tabular}
    }
    \caption{Top performing prompts for each cluster of Parl-Multi task (using PaLM 2-L-IT for \llmopt\:and text-bison for \llmscore).}
    \label{tab:parl_multi_2}
\end{table}

\noindent \textbf{Results with text-unicorn as both scorer and optimizer}
\begin{table}[ht]
\renewcommand{\arraystretch}{1.2}
\scriptsize
\begin{tabular}{p{0.06\textwidth}p{0.042\textwidth}p{0.048\textwidth}p{0.04\textwidth}p{0.04\textwidth}p{0.02\textwidth}p{0.04\textwidth}}
\toprule
\textbf{Method} & \textbf{Parl-S} & \textbf{Parl-M} & \textbf{OpenAI} & \textbf{ETHOS} & \textbf{Liar} & \textbf{Sarcasm} \\
\hline
\multicolumn{7}{c}{Prompt Optimizers} \\
CE    & 0.78  & \textbf{0.84} & 0.85 & \textbf{0.86} & \textbf{0.74} & 0.65 \\
ProTeGi        & 0.75   & 0.65 & 0.83 & 0.88 & 0.71 & \textbf{0.73} \\
Prompt-Breeder       & \vnudge{0.44}   & \vnudge{0.34} & \vnudge{0.77} & \vnudge{0.83} & \vnudge{0.71} & \vnudge{0.22}\\
\hline
\multicolumn{7}{c}{Prompt Optimizers + MoE} \\
CE & \textbf{0.79}  & 0.78 & \textbf{0.87} & 0.85 & \textbf{0.74} & 0.65  \\
\hline
\multicolumn{7}{c}{Standard Prompting Techniques} \\
Zero-shot    & 0.76   & 0.67 & 0.74 & 0.76 & 0.68 & 0.56 \\
Few-shot ($n$=8)         & \vnudge{0.75}   & \vnudge{0.72} & \vnudge{0.80} & \vnudge{0.78} & \vnudge{0.74} & \vnudge{0.59}\\
Chain of Thought       & \vnudge{0.71}   & \vnudge{0.48} & \vnudge{0.77} & \vnudge{0.73} & \vnudge{0.46} & \vnudge{0.39}\\
\hline
\bottomrule
\end{tabular}
\caption{Main results from the evaluation when using `text-unicorn' for scoring. Values are F1 scores, averaged over three runs for ConstitutionalExperts (CE), ProTeGi, and PromptBreeder.}
\label{tab:main-res-text-unicorn}
\end{table}

\noindent \textbf{Sample prompt templates}
\label{sec:scoring_model_template}
\noindent Below is the prompt template used for classification.
{\small
\begin{verbatim}
Consider the following example:
{% for input_feature in input_features %}
    <{{input_feature.name}}>
        {{input_feature.value}}
    </{{input_feature.name}}>
{% endfor %}
{{task_description}} Let's think step-by-step.
Consider the following possible answers:
{% for class in classes -%}
answer_{{class.id}}: 
{% for attribute in class.attributes -%}
    {{attribute}}
{% endfor -%}
{% endfor -%}

Provide the answer that best applies to this example: 
answer_
\end{verbatim}
}

The class "attributes" in the prompt template are the parameters to be optimized during training. 

\begin{table}[ht]
\begin{tabular}{lll}
                                 & n\_experts=2               & n\_experts=3               \\ \cline{2-3} 
\multicolumn{1}{l|}{Parl Single} & \multicolumn{1}{l|}{0.031} & \multicolumn{1}{l|}{0.021} \\ \cline{2-3} 
\multicolumn{1}{l|}{Parl Multi}  & \multicolumn{1}{l|}{0.059} & \multicolumn{1}{l|}{0.016} \\ \cline{2-3} 
\multicolumn{1}{l|}{OpenAI}      & \multicolumn{1}{l|}{0.014} & \multicolumn{1}{l|}{0.022} \\ \cline{2-3} 
\multicolumn{1}{l|}{ETHOS}       & \multicolumn{1}{l|}{0.028} & \multicolumn{1}{l|}{0.024} \\ \cline{2-3} 
\multicolumn{1}{l|}{Liar}        & \multicolumn{1}{l|}{0.024} & \multicolumn{1}{l|}{0.020} \\ \cline{2-3} 
\multicolumn{1}{l|}{Sarcasm}     & \multicolumn{1}{l|}{0.174} & \multicolumn{1}{l|}{0.203} \\ \cline{2-3} 
\end{tabular}
\caption{Silhouette scores for datasets.}
    \label{tab:silouhette}
\end{table}

\begin{table}[ht]
\begin{tabular}{ll}
                                 & Initial prompt                                              \\ \cline{2-2} 
\multicolumn{1}{l|}{Parl Single} & \multicolumn{1}{l|}{Does the example contain toxic speech?} \\ \cline{2-2} 
\multicolumn{1}{l|}{Parl Multi}  & \multicolumn{1}{l|}{Does the example contain toxic speech?} \\ \cline{2-2} 
\multicolumn{1}{l|}{OpenAI}      & \multicolumn{1}{l|}{Does the example contain toxic speech?} \\ \cline{2-2} 
\multicolumn{1}{l|}{ETHOS}       & \multicolumn{1}{l|}{Does the example contain toxic speech?} \\ \cline{2-2} 
\multicolumn{1}{l|}{Liar}        & \multicolumn{1}{l|}{Is the statement true?}                 \\ \cline{2-2} 
\multicolumn{1}{l|}{Sarcasm}     & \multicolumn{1}{l|}{Is the tweet sarcastic?}                \\ \cline{2-2} 
\end{tabular}
\caption{Initial prompts for datasets.}
    \label{tab:initial_prompts}
\end{table}

\end{document}